\documentclass[11pt,letterpaper]{article}

\usepackage[margin=1in]{geometry}
\usepackage{graphicx}
\usepackage{dcolumn}
\usepackage{bm}

\usepackage[utf8]{inputenc}
\usepackage[T1]{fontenc}
\usepackage{mathptmx}
\usepackage{etoolbox}

\usepackage{xcolor}
\usepackage{siunitx}
\usepackage{hyperref}
\hypersetup{
  colorlinks=true,
  linkcolor=blue,
  citecolor=blue,
  urlcolor=blue
}

\usepackage{tikz}
\usetikzlibrary{arrows.meta,positioning,shapes.geometric,shapes.symbols}

\newcommand{\figref}[1]{Figure~\ref{#1}}

\begin{document}

\title{From Natural Language to Control Signals: A Conceptual Framework for Semantic Channel Finding in Complex Experimental Infrastructure}

\author{
Thorsten Hellert$^{1,*}$,
Nikolay Agladze$^2$,
Alex Giovannone$^2$,
Jan Jug$^5$,
Frank Mayet$^3$,
Mark Sherwin$^2$, \\
Antonin Sulc$^1$,
Chris Tennant$^4$ \\[1em]
\small $^1$Lawrence Berkeley National Laboratory, Berkeley, California 94720, USA \\
\small $^2$University of California Santa Barbara, Santa Barbara, California 93106, USA \\
\small $^3$Deutsches Elektronen-Synchrotron DESY, Notkestrase 85, 22607 Hamburg, Germany \\
\small $^4$Thomas Jefferson National Accelerator Facility, Newport News, Virginia 23606, USA \\
\small $^5$ Cosylab USA, Menlo Park, California 94025, USA \\[0.5em]
\small $^*$Corresponding author: \texttt{thellert@lbl.gov}
}

\date{}

\maketitle


\begin{abstract}
    Modern experimental platforms such as particle accelerators, fusion devices, telescopes, and industrial process control systems expose tens to hundreds of thousands of control and diagnostic channels accumulated over decades of evolution. Operators and AI systems rely on informal expert knowledge, inconsistent naming conventions, and fragmented documentation to locate signals for monitoring, troubleshooting, and automated control, creating a persistent bottleneck for reliability, scalability, and language-model–driven interfaces. We formalize semantic channel finding-mapping natural-language intent to concrete control-system signals-as a general problem in complex experimental infrastructure, and introduce a four-paradigm framework to guide architecture selection across facility-specific data regimes. The paradigms span (i) direct in-context lookup over curated channel dictionaries, (ii) constrained hierarchical navigation through structured trees, (iii) interactive agent exploration using iterative reasoning and tool-based database queries, and (iv) ontology-grounded semantic search that decouples channel meaning from facility-specific naming conventions. We demonstrate each paradigm through proof-of-concept implementations at four operational facilities spanning two orders of magnitude in scale-from compact free-electron lasers to large synchrotron light sources-and diverse control-system architectures, from clean hierarchies to legacy environments. These implementations achieve 90–97\% accuracy on expert-curated operational queries.\end{abstract}

\section{Introduction}
\label{sec:intro}

\noindent Complex experimental and industrial facilities—from particle accelerators and fusion devices to telescopes, chemical plants, and power grids—are among the most intricate operational environments in modern science and engineering.
A single facility may expose tens to hundreds of thousands of control and diagnostic channels accumulated over decades of hardware evolution, subsystem upgrades, and vendor integrations.
High-level operational tasks such as stabilizing beam parameters, recovering from faults, optimizing process efficiency, or preparing custom configurations, ultimately require identifying and interacting with the correct process variables (PVs) or control signals under strict safety, performance, and real-time constraints.
Even when operators know the physical quantity of interest, translating that intent into the relevant channel names is often nontrivial, facility-specific, and dependent on institutional knowledge.

Today this mapping relies heavily on informal expert knowledge, inconsistent naming conventions, and fragmented documentation.
Legacy channels often encode subsystem structure in cryptic, fixed-length identifiers shaped by decades-old software constraints, while newer subsystems introduce more descriptive conventions but rarely update older names for backward compatibility.
Different subsystems like magnets, diagnostics, vacuum, RF or instrumentation, develop separate terminologies and ontologies, and descriptions are scattered across configuration files, wikis, personal notes, and undocumented tribal knowledge.
As a result, knowledge of the channel space is implicitly distributed across specialists, creating steep learning curves for new operators and persistent bottlenecks for troubleshooting under time pressure.

These structural issues create persistent challenges.
New operators face steep learning curves; troubleshooting requires rapid channel identification under time pressure; and documentation rarely keeps pace with hardware evolution.
For emerging AI-based control assistants, the bottleneck is even more acute.
Language-model–driven systems cannot safely inspect machine state, retrieve history, or execute control actions unless they can reliably map natural-language intent to concrete control-system channels.

Large language models (LLMs) now enable a new interaction layer in which operators and autonomous agents specify goals directly in natural language.
Within this layer, \emph{semantic channel finding} becomes a foundational capability: given a free-form description (for example, “orbit corrector strengths in the booster” or “temperatures near the undulator chamber”), the system must identify the corresponding PVs and related readback, setpoint, or status channels needed for safe operation.

However, control systems vary widely in scale, naming conventions, and data quality, making a single universal retrieval architecture infeasible.
In this work, we formalize semantic channel finding as a general architectural problem in complex experimental and industrial infrastructure; applicable not only to accelerators but also to fusion devices, telescope arrays, chemical process control, power systems, and any large-scale facility where natural-language interfaces to control systems are desired.
Rather than proposing a single universal solution, we present a conceptual framework that systematically organizes the design space into four paradigms, each optimized for different facility-specific \emph{data regimes} characterized by control-system scale, naming convention quality, presence of hierarchical structure, and availability of semantic metadata.
Paradigm~1 (\emph{direct in-context lookup}) is well suited to small or curated channel sets (hundreds to low thousands) where a complete dictionary fits comfortably in modern LLM context windows, enabling zero-shot semantic matching without external retrieval.
Paradigm~2 (\emph{constrained hierarchical navigation}) applies when channel organization admits a strict tree structure: agents traverse pre-defined levels with strongly typed choices at each step, eliminating hallucination while keeping computational cost linear in tree depth rather than channel count.
Paradigm~3 (\emph{interactive agent exploration}) handles large databases where structure must be discovered dynamically: ReAct-style agents reason iteratively about what information they need, call database tools to explore compositional naming schemes or middle-layer abstractions, and refine their search based on observations.
Paradigm~4 (\emph{ontology-based semantic search}) elevates reasoning to the level of device types and signal semantics, grounding devices, signals, and relationships in a shared domain model to enable portable, facility-agnostic queries at the cost of upfront ontology development and mapping effort.
The framework provides actionable guidance: facilities can select paradigms based on their existing infrastructure and data quality rather than undertaking extensive architectural redesign.

Building on this perspective, our contributions are threefold.
\textbf{First}, we formalize semantic channel finding as a general architectural problem and develop a four-paradigm conceptual framework (Sections~\ref{sec:direct_lookup}, \ref{sec:hierarchical}, \ref{sec:interactive}, and \ref{sec:ontology}) that enables systematic selection of appropriate architectures based on facility-specific data regimes—control-system scale, naming convention quality, hierarchical structure, and available metadata.
This taxonomy provides actionable guidance for practitioners facing the channel-finding problem across diverse scientific and industrial contexts.
\textbf{Second}, we validate the practical feasibility of each paradigm through proof-of-concept implementations at operational facilities spanning two orders of magnitude in scale and operating under fundamentally different control-system architectures: the compact Terahertz Free Electron Laser at UC Santa Barbara~\cite{Ramian1992UCSB} (300 channels, flat organization), the European XFEL~\cite{altarelli2015european, Decking2020} at Deutsche Elektronen-Synchrotron DESY (compositional DOOCS naming), the Advanced Light Source (ALS)~\cite{Hellert2024zrj} at Lawrence Berkeley National Laboratory (legacy heterogeneous conventions mediated through middle-layer abstraction), and CEBAF~\cite{CEBAF}/ALS ontology-based deployments at Jefferson Lab.
Each implementation is validated on expert-curated operational queries representative of real control-room tasks, demonstrating 90–97\% accuracy and establishing that the framework's paradigms are deployable under real-world constraints.
\textbf{Third}, we provide open-source, plug-and-play implementations of the first three paradigms (direct lookup, hierarchical navigation, and middle-layer exploration) within the Osprey framework~\cite{hellert2025osprey,osprey2025}, including database-generation tools, interactive interfaces, and minimal-configuration tutorials to support adoption at other facilities without requiring custom infrastructure development.
Section~\ref{sec:future} discusses potential extensions such as hybrid pipelines, reinforcement-learning-enhanced navigation, dynamic agent memory, and cross-facility knowledge transfer built on shared ontologies.

\section{Paradigm 1: Direct Lookup}\label{sec:direct_lookup}

In control systems where the number of channels of interest is below about 1000, the most straightforward path to semantic channel finding is to treat a large language model itself as a search engine over a complete channel dictionary.
Instead of asking the model to discover channels by navigating tools or databases, we provide it with a carefully formatted list of all available channels and their natural language descriptions directly in its context window.
Operator queries are then posed as ordinary natural-language prompts.
Crucially, this does not require the entire control system to be small: operators can define a focused subset of a few hundred high-value channels from a much larger machine and expose only that subset to the model.
The model’s in-context learning capabilities then allow it to map between free-form descriptions of desired quantities and the most semantically compatible entries in this dictionary without any additional training, in a way that closely matches how operators already think about and describe the machine.

This direct lookup paradigm exploits the fact that modern language models can process thousands of tokens and perform robust semantic matching within a single prompt.
When the full channel inventory fits comfortably into the context window, this yields a practical “zero-setup” solution: no external retriever, vector index, or database infrastructure is required, updates reduce to regenerating the prompt text, and all reasoning remains transparent in a single model call.
At the same time, the approach is inherently bounded by context length and prompt quality, which makes it most suitable for small to medium-scale facilities with descriptive names and stable channel sets.

In this section we make this paradigm concrete in the setting of the UCSB Free Electron Laser (FEL).
We describe how to construct an in-context channel dictionary that is optimized for language-model consumption, how to structure the query pipeline so that multi-target questions can be decomposed and matched reliably, and how to validate and iteratively correct the model’s outputs against the ground-truth database.
This case study illustrates both the strengths and limitations of direct lookup and motivates the more scalable hierarchical and ontology-based approaches developed in the subsequent sections.

\subsection{Case Study 1: In-Context Semantic Search}\label{subsec:ucsb_fel}

The UCSB FEL is a compact THz laser facility with approximately 300 control system channels spanning electron source, acceleration, beam transport, diagnostics, undulator, and energy-recovery subsystems.
The control system has evolved under a flat organizational structure with no systematic naming conventions and highly cryptic address names, making it a challenging but representative testbed for semantic channel finding.

To address this, we construct a two-layer channel dictionary in which human-readable \texttt{channel} names (for example, \texttt{TerminalVoltageSetPoint}) are systematically mapped to underlying control system \texttt{address} identifiers (such as \texttt{TMVST}).
Descriptive names are chosen so that both the name and the accompanying description reinforce the same semantic concepts, improving the robustness of language-model matching.
An automated name-generation procedure uses an LLM to extract salient semantic components from legacy descriptions and assemble consistent PascalCase channel names.
A small set of 58 template entries is then expanded programmatically to a full database of roughly 300 channels, which greatly simplifies maintenance.

\begin{figure}
    \centering
    \includegraphics[width=1\linewidth]{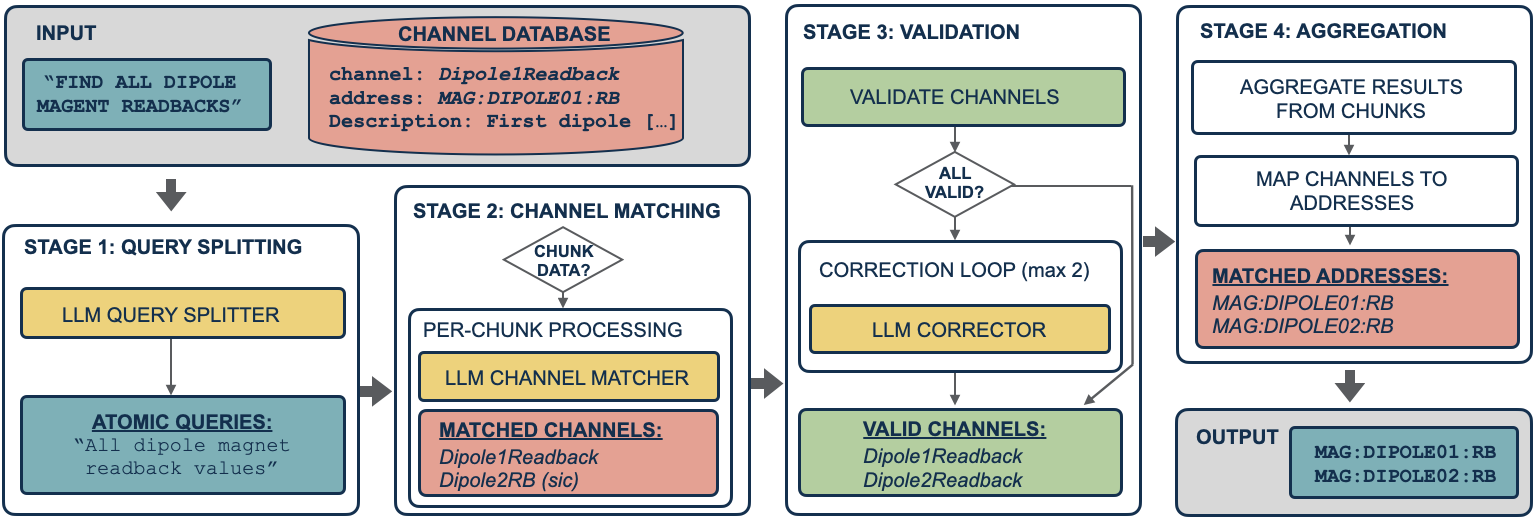}
    \caption{Four-stage pipeline for direct in-context semantic channel finding. Stage~1 decomposes multi-target queries into atomic sub-queries using structured outputs. Stage~2 performs semantic matching by providing the complete channel database directly in the LLM context, with precision-oriented tuning to minimize false positives. Stage~3 validates candidate channels against the ground-truth database and applies iterative correction to eliminate invalid or non-existent channel suggestions.}
    \label{fig:incontext}
\end{figure}

Operator queries are handled by a three-stage pipeline with explicit validation, as depicted in \figref{fig:incontext}.
In Stage~1 (query decomposition), multi-target requests such as ``pressure and temperature'' are split into atomic sub-queries using structured outputs from the model.
In Stage~2 (semantic matching), the complete channel database is provided directly in the LLM context together with facility-specific terminology, and the system is tuned for precision over recall so as to minimize false positive channel suggestions.
In Stage~3 (validation), all candidate channels are verified against the ground-truth database, and iterative correction reduces the rate of invalid or non-existent channels.

On a validation set of 30 expert-curated operational queries, this in-context approach achieves over 90\% exact-match accuracy using Claude 4.5 Haiku while maintaining an end-to-end latency of only 2--3 seconds per query.
The resulting system is cost-effective in routine operation, with negligible cloud API expenditures relative to typical facility computing costs.

\textbf{Generalization:} The in-context lookup approach is control-system agnostic and readily adaptable to any facility with a manageable number of channels or a well-defined subset of high-value channels.
Facilities wishing to implement this method at their own site can build on the open-source Osprey framework, which includes a complete reference implementation in the Control Assistant Tutorial~\cite{osprey_tutorial_channel_finder}.
Deployment requires a channel database with natural-language descriptions, but the pipeline itself is configured entirely via YAML, with no code changes needed to switch between in-context, hierarchical, and middle-layer modes (see Sections \ref{sec:hierarchical} and \ref{sec:interactive}).
The framework includes validation tools, an interactive command-line interface for rapid testing, and a validation suite for systematic evaluation.
The UCSB FEL implementation, encompassing roughly 300 channels and 30 validation queries with over 90\% accuracy using Claude Haiku 4.5, is provided as a worked example to facilitate adoption.
When curated channel dictionaries no longer fit comfortably into a single model context, we turn next to hierarchical navigation over large channel sets.

\section{Paradigm 2: Hierarchical Navigation}\label{sec:hierarchical}

While direct in-context lookup works well when a curated channel dictionary fits entirely within the model's context window, many modern control systems expose tens to hundreds of thousands of channels, far beyond what can be reliably presented in a single prompt.
When these large channel sets admit a strict hierarchical organization, for example a fixed tree structure with well-defined levels such as system, family, device, field, and subfield, a deterministic descent strategy becomes viable.

The hierarchical navigation paradigm exploits this structure by constraining the language model at each step to select only from valid options at that level of the hierarchy using structured outputs with dynamically constructed enumeration types.
This eliminates hallucination by construction: the model cannot generate invalid channel names or non-existent paths.
The approach requires that channels admit hierarchical decomposition (typically system $\to$ family $\to$ device $\to$ field $\to$ subfield), but is otherwise control-system agnostic.

The overall architecture proceeds in three stages.
Stage~1 (query splitting) decomposes complex multi-target queries into atomic sub-queries using structured outputs.
Stage~2 (recursive hierarchical navigation) is the core: the agent navigates through hierarchy levels in sequence, selecting from dynamically constrained option sets at each level.
When multiple options are selected at semantically meaningful branch points, navigation recursively branches with each path independently exploring remaining levels.
Stage~3 (channel assembly) combines selections using a facility-specific naming-pattern template and validates results against the ground-truth database.

Computational cost scales linearly with hierarchy depth (typically 4--6 LLM calls) rather than total channel count, enabling the method to handle databases too large for any context window.
The trade-off is higher per-query latency due to sequential LLM calls, but in exchange the approach offers essentially unbounded scale.

In this section we demonstrate the practical deployment of hierarchical navigation through a production implementation and show how to construct hierarchical databases with natural-language descriptions at each level, design recursive query pipelines that branch only at semantically meaningful choice points, and assemble and validate final channel identifiers from hierarchical selections.

\subsection{Case Study 2: Production EPICS Deployment}\label{subsec:hierarchical}

\begin{figure}
    \centering
    \includegraphics[width=1\linewidth]{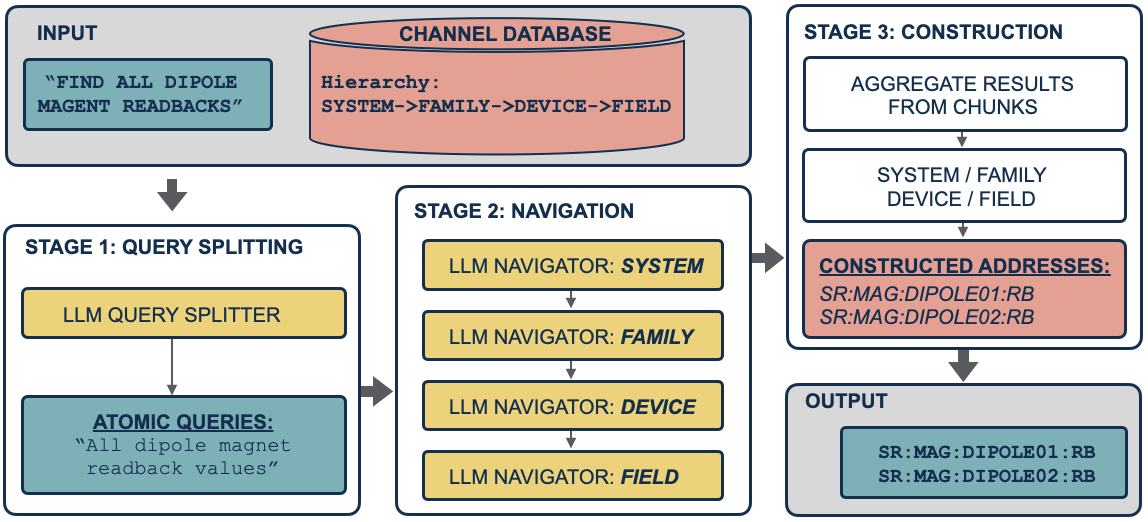}
    \caption{Hierarchical navigation pipeline for structured channel finding. Stage~1 decomposes complex queries into atomic sub-queries. Stage~2 performs recursive hierarchical navigation through control system levels (system $\to$ subsystem $\to$ device $\to$ subdevice $\to$ signal $\to$ suffix), selecting from dynamically constrained option sets at each level. When multiple options are selected at branch points, navigation recursively explores each path independently. Stage~3 assembles final channel identifiers from hierarchical selections using facility-specific naming patterns and validates results against the ground-truth database.}
    \label{fig:hierachical}
\end{figure}

To illustrate the practical deployment of hierarchical channel finding, we present a production implementation in EPICS, recently developed by Cosylab as part of a modern accelerator commissioning project.

As is common in EPICS implementations, the actual PVs are instantiated via macro expansions, so the final number of PVs is dependant on the accelerator architecture and the number of devices used.
For the purposes of testing the channel finder, we created an IOC that manages 1,230 process variables spanning the four major subsystems: vacuum (72\% of PVs), RF (18\%), magnets (6\%), and diagnostics (3\%).
The IOC is using real hardware devices and interfaces to support realistic environment and PV names.

The control system employs a contemporary naming convention designed from the ground up to support both human operators and automated tooling.
The PV names created via this naming convention are unique and provide structural information about the device, equipment, or signal.

The channel database implements a six-level hierarchy: system $\to$ subsystem $\to$ device $\to$ subdevice $\to$ signal $\to$ suffix.
This structure mirrors the physical control system architecture and accommodates the heterogeneous device integration typical of production facilities.

For example, the vacuum subsystem contains both multi-channel gauge controllers - each managing ion gauges, thermal gauges, and automated setpoint controls - and ion pump controllers with high-voltage supplies and pressure monitoring.
Despite different signal structures and channel counts per device type, the unified hierarchical representation enables consistent navigation across all device classes.

A key feature of this modern design is the handling of optional hierarchy levels.
Channels exist both at the device level (firmware version, unit settings) and subdevice level (individual gauge pressures, pump voltages).
The navigation pipeline accommodates queries that may terminate at any hierarchy depth without forcing traversal through all levels - a deliberate design choice that reflects best practices in contemporary control system architecture.
This flexibility is essential for matching the semantic granularity of operator queries: a request for ``magnet power supply firmware version'' should not require artificially descending through subdevice and signal levels when the relevant channel exists directly at the device level.

The database exploits range expansion to achieve compact representation of large channel spaces.
A single device definition with subdevice range specifications - for instance, 4 ion gauges per controller, 8 pressure setpoints per controller, 6 pump channels per controller - generates 145--152 PVs per vacuum device instance through Cartesian product assembly.
This approach balances realism with maintainability: list-based expansion handles heterogeneous devices (ion and thermal gauges for gauge controllers), while range-based expansion covers identical units, such as magnet power supplies.
The result is a natural representation of actual facility topology that can evolve incrementally as new devices are commissioned.

The suffix level encodes EPICS control system conventions that create 2--3$\times$ signal multiplicity for controllable parameters such as setpoint (SP), readback (RB) or command (CMD).
Semantic understanding of read versus write versus measured channels is essential for correct query resolution.
An operator request to ``set vacuum setpoint'' must resolve to the SP suffix, while ``check current vacuum pressure'' requires the RB suffix.

Per-query latency remains constant at 4--6 LLM calls for the six-level hierarchy, independent of whether the database contains 1,000 or 10,000+ channels.
This validates hierarchical navigation as a viable approach for facilities planning control system expansion without redesigning channel finding infrastructure.

Database maintenance patterns match operational evolution: adding a new device instance (e.g., a fourth vacuum gauge controller) requires a single line in the expansion list; adding a new device type within an existing subsystem requires new signal definitions with metadata; adding a new subsystem requires a new branch in the tree structure.
These incremental updates align naturally with actual facility commissioning workflows, enabling the channel finding system to grow alongside the control system it serves.

\textbf{Generalization:} The hierarchical tree navigation approach is control-system agnostic and readily adaptable to any facility whose channels admit hierarchical organization.
Facilities wishing to implement this method at their own site can build on the open-source Osprey framework, which includes a complete plug-and-play reference implementation in the Control Assistant Tutorial~\cite{osprey_tutorial_channel_finder} and a working example database spanning 1,050~channels across four systems.
The framework includes validation tools, an interactive command-line interface for rapid testing, and a benchmark suite for systematic evaluation.
The framework supports advanced features demonstrated in the Cosylab deployment, including optional hierarchy levels, range-based device expansion, and flexible separator patterns, enabling deployment across diverse naming conventions without framework modification.

\section{Paradigm 3: Interactive Agent Exploration}\label{sec:interactive}

When channel databases are large but lack the strict hierarchical structure required for deterministic tree navigation, or when control systems support flexible querying over compositional naming schemes or searchable metadata, an alternative paradigm becomes viable: interactive exploration through iterative reasoning.
Rather than descending a fixed tree level by level, agents in this paradigm reason explicitly about what information they need, invoke tools to query the database or control system, observe the results, and adaptively refine their search strategy based on what they discover.

This interactive approach is built on the ReAct framework~\cite{yao2023react}, which interleaves natural-language reasoning with tool-based actions in a loop: the agent decides what to do next, executes a tool call, receives an observation, and uses that feedback to inform subsequent steps.
For semantic channel finding, this means the agent can list available device types, filter by location or property, inspect intermediate candidates, and iteratively narrow the search space until target channels are identified.
The resulting exploration is more flexible than constrained tree navigation: it can handle irregular structure, adapt to unexpected database organization, and exploit facility-specific naming conventions, but at the cost of higher computational overhead and the need for careful tool design to maintain safety and prevent runaway query patterns.

The effectiveness of interactive exploration depends critically on two factors: the quality of the underlying data (whether names and descriptions are informative enough to guide reasoning), and the design of the tool interfaces through which the agent interacts with the control system.
When these factors align, ReAct-style agents can achieve high accuracy even in challenging environments with decades of heterogeneous legacy conventions.
When data quality is poor—sparse descriptions, cryptic names, inconsistent terminology—the approach may require an intermediate abstraction layer to provide the semantic clarity necessary for reliable reasoning.

In the remainder of this section, we examine three distinct instantiations of interactive agent exploration.
We begin with a system built around compositional DOOCS-style naming, where systematic address structure enables reasoning-based navigation augmented by fuzzy search (Case Study 3).
We then analyze an instructive failure case where direct database queries over poor-quality metadata led to inefficient shotgun search patterns, illustrating the data quality prerequisites for this paradigm (Case Study 4).
Finally, we show how targeting a middle-layer abstraction overcame these fundamental challenges at a legacy facility, enabling production deployment with strong accuracy and efficiency (Case Study 5).
Together, these case studies reveal when interactive exploration succeeds, when it fails, and how to design robust implementations across diverse control system architectures.

\subsection{Case Study 3: Compositional Address Navigation}\label{subsec:react_compositional}

We begin by examining an approach for control systems with compositional naming schemes.
This case study develops a PV finder for DOOCS-like control systems~\cite{DOOCS}, where addresses follow a structured pattern \texttt{FACILITY/DEVICE/LOCATION/PROPERTY} with accompanying property descriptions.
Such a system is, for example, deployed at Deutsches Elektronen-Synchrotron DESY at the \SI{3}{\km} long European XFEL.
The approach exploits systematic naming conventions to enable iterative refinement through tool-assisted reasoning, providing an alternative to tree navigation when channel spaces exhibit regular compositional structure rather than strict hierarchies.

The core requirement is that channel addresses decompose into semantically meaningful components.
In DOOCS, a \emph{facility} combines accelerator and component group names (e.g., \texttt{XFEL.DIAG} for XFEL diagnostics), a \emph{device} specifies the device class (e.g., \texttt{CAMERA}, \texttt{BPM}), a \emph{location} identifies the specific installation point, and a \emph{property} names the accessible parameter with a short description (e.g., \texttt{IMAGE The camera image (8 bit)}).
This compositional structure, combined with descriptive component names, enables reasoning-based navigation without requiring a pre-structured hierarchy.

The architecture employs the open-weights reasoning model \texttt{gpt-oss:120b}~\cite{openai2025gptoss120bgptoss20bmodel} equipped with control-system-specific tools.
The tool suite enables querying each address component level with optional filtering: \texttt{list\_facilities}, \texttt{list\_devices}, \texttt{list\_locations}, \texttt{list\_properties}, and \texttt{compose\_address} for final assembly and validation with the control system.
The system prompt provides control system basics, a short glossary, and a clear work plan: list facilities, reason about selection, proceed to devices, then locations, and finally properties.
Property descriptions guide final selection before address composition.
Invalid addresses trigger re-reasoning within iteration limits, eliminating hallucination while maintaining systematic exploration.

Performance optimization introduces a hybrid approach combining reasoning with fuzzy search.
The \texttt{guess\_addresses} tool performs top-k fuzzy search across the DOOCS hierarchy, ranking validated addresses by property-centric scoring with additive boosts from facility, device, and location matches.
The modified workflow first identifies relevant facilities, then attempts rapid retrieval via fuzzy search using model-determined query hints.
If the top-k candidates (default $k=5$) satisfy the query, they are returned immediately.
Otherwise, the agent falls back to systematic component-wise construction.
This hybrid strategy reduces retrieval time while maintaining accuracy through internal reasoning validation.

Validation on 60 expert-curated queries spanning multiple accelerators achieves 96\% accuracy using \texttt{gpt-oss:120b} with its reasoning effort set to medium and zero temperature.
The system correctly retrieved 58 of 60 addresses, with ground truth occasionally containing multiple valid answers for queries targeting redundant services.
The average reasoning depth of $8\pm4$ round-trips demonstrates efficient convergence while maintaining robustness against invalid selections.

The scalability profile differs from hierarchical tree navigation in several key dimensions.
While tree navigation requires pre-structured hierarchies with descriptions at each level, ReAct agents operate directly on flat component lists with compositional naming.
Computational cost scales with reasoning depth and component list sizes at each level, but the fuzzy search optimization provides sub-linear scaling for common queries.
The approach excels when naming conventions are systematic and descriptive, enabling semantic reasoning about component relationships without explicit hierarchy encoding.

\textbf{Generalization:} The ReAct agent approach with compositional naming is adaptable to any control system with structured address patterns and descriptive component names.
Implementation requires: (1) a decomposable address format with semantically meaningful components, (2) query tools for each component level with optional filtering, (3) an address composition and validation tool, and (4) a system prompt encoding the control system structure and navigation strategy.
Facilities can adapt the reference implementation by modifying the tool implementations to match their address structure, updating the system prompt with facility-specific terminology, and optionally tuning the fuzzy search scoring weights.
The approach is particularly effective for control systems with regular, descriptive naming schemes where component semantics are encoded in the names themselves rather than requiring external documentation.

\subsection{Case Study 4: Direct Database Queries}\label{subsec:react_database}

Before describing the successful middle-layer-based implementation at ALS, we examine an alternative approach that works well when certain data quality prerequisites are met, but fails instructively when they are not.
Many facilities maintain searchable channel finder databases with metadata fields that can be queried using regex patterns and filters—essentially SQL-like endpoints over the control system namespace \cite{shen2010nsls2_commissioning} \cite{shroff2023channelfinder}.
This naturally suggests equipping a ReAct agent with database query tools to perform free-form exploration, leveraging language models' strong out-of-the-box capabilities with SQL and related structured query languages.

In this approach, the agent is given tools that issue queries directly to the channel finder database, combining regular-expression searches with metadata filters.
The language model constructs and refines these queries iteratively: it inspects intermediate results, adjusts patterns or filters, and launches follow-up queries as needed.
Because the agent operates directly on the existing database, the method does not require any custom preprocessing or explicit construction of control-system hierarchies, and can be deployed with minimal additional infrastructure.

The effectiveness of this strategy, however, depends critically on data quality.
For the agent to reliably home in on relevant process variables (PVs), either PV names must follow consistent, self-descriptive conventions that can be parsed semantically by the LLM, or the database must contain comprehensive, accurate description fields for each channel.
In practice, the best performance is obtained when both conditions are satisfied: systematic naming schemes paired with reliable, up-to-date descriptive metadata.

When these prerequisites are met, the ReAct-plus-database approach can perform quite well.
With a small number of carefully designed few-shot examples, the model learns to synthesize SQL-like queries that retrieve relevant subsets of channels with high precision.
Direct database access obviates the need for precomputed hierarchical structures or manually curated mappings, and the reuse of existing channel finder infrastructure keeps deployment overhead low.

Our initial implementation at the Advanced Light Source (ALS) illustrates the limitations of this strategy in the presence of poor data quality.
The ALS control system, shaped by over 40 years of operational history, exhibits inconsistent naming conventions across subsystems and sparse, often outdated description fields—lacking both of the data quality prerequisites identified above.
In this setting, the agent is forced into broad, exploratory query patterns—an inefficient "shotgun" search behavior that issues many overlapping queries in an attempt to reconstruct the underlying semantics.
This behavior leads to substantial resource costs (on the order of \$1 per user query) and end-to-end latencies of 10–20 seconds, driven by multiple rounds of query generation, execution, and refinement.
Although the final answers were often reasonable, these efficiency and robustness issues rendered the approach unsuitable for production use at ALS.

\textbf{Generalization:} These observations have two main implications.
First, the ReAct-plus-database pattern remains attractive for facilities with disciplined PV naming conventions or well-maintained metadata, where it can provide accurate results with modest engineering effort.
Second, for legacy systems with heterogeneous, weakly structured naming and sparse or unreliable descriptions, an additional abstraction layer is essential.
Recognition of these data-quality-driven limitations directly motivated the development of the MML-based approach described in the next section, in which the hierarchy is imposed explicitly rather than inferred opportunistically from noisy channel finder metadata.

\subsection{Case Study 5: Middle-Layer Abstraction Navigation}\label{subsec:react_abstraction}

Having established that direct database querying fails when naming conventions and descriptions are inadequate, we now describe how the ALS overcame these fundamental data quality challenges.
Rather than attempt to impose structure on the heterogeneous raw control system namespace, the ALS implementation~\cite{hellert2025agenticaimultistagephysics} builds semantic channel finding on top of an existing middle-layer abstraction—the MATLAB Middle Layer (MML)—which organizes approximately 10,000 key process variables into a consistent tree structure spanning six accelerator systems (storage ring, booster, transfer lines, and injector components) \cite{corbett2002mml}.

The key architectural decision is to target the MML abstraction rather than the raw control system.
As demonstrated by the failure of the SQL-based approach, the native ALS control system lacks the naming consistency and descriptive metadata necessary for reliable LLM-based navigation.
The DESY approach described earlier similarly would not be viable at ALS: that method's reliance on systematic, hierarchical naming conventions (which enable both fuzzy search and the \texttt{guess\_addresses} tool) assumes a level of naming discipline that does not exist in the 40-year legacy ALS namespace.
The MML Accelerator Object, however, provides a clean abstraction layer that organizes heterogeneous raw process variables into a consistent tree structure with normalized device families and standardized field names.
This architectural choice transforms what would otherwise be an intractable data quality problem into a well-structured hierarchical navigation task amenable to standard agent-based exploration techniques.

\begin{figure}[ht]
    \centering
    \includegraphics[width=1\linewidth]{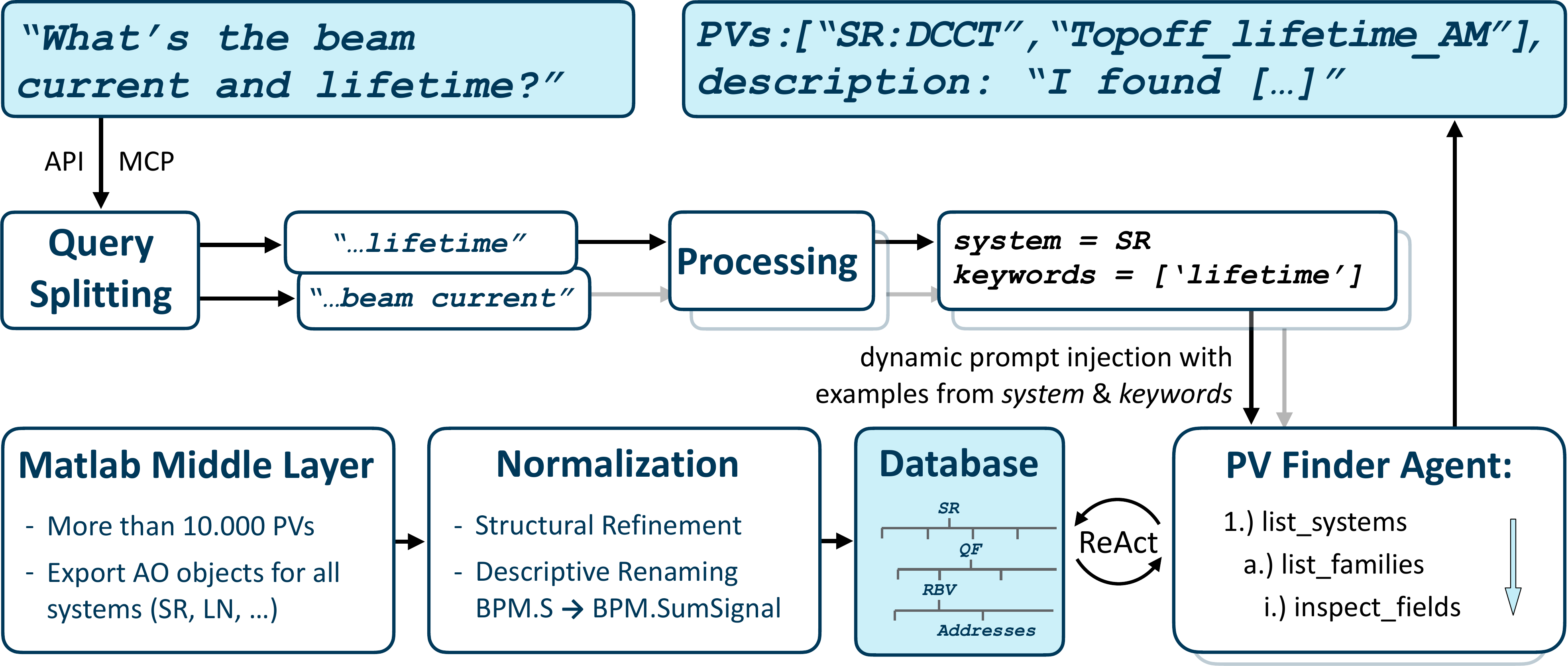}
    \caption{Schematic of the PV Finder implementation. User queries undergo decomposition and semantic analysis to extract accelerator systems and keywords. Domain-specialized agents then navigate the normalized MML Accelerator Object representation through bounded tool interfaces to identify target EPICS process variables.}
    \label{fig:pv_finder}
\end{figure}

The ALS implementation employs a three-stage pipeline that combines query preprocessing with dynamic, domain-specific agent construction (Figure~\ref{fig:pv_finder}).
In the first stage, incoming user queries undergo analysis and splitting: queries that target multiple independent systems or device types are decomposed into constituent sub-queries.
Simultaneously, semantic keyword detection analyzes the query content to determine which accelerator systems and device types are relevant.
This preprocessing step is then used to construct specialized ReAct agent prompts at runtime.
The system maintains an extensive library of query-tool-result patterns, categorized by accelerator system (storage ring, booster, transfer lines, etc.) and query type.
Based on the keyword analysis, only the examples relevant to the detected domain are included in the agent prompt.
Rather than deploying a single general-purpose agent, the system dynamically constructs domain-specialized agents tailored to each query: an agent handling storage ring beam position monitor queries receives different navigation examples than one addressing booster magnet queries.
This example-driven approach teaches subsystem-specific navigation patterns without requiring explicit fine-tuning or model training.

In the second stage, the sub-queries are processed in parallel, with each invoking its domain-specialized ReAct agent.
Each agent navigates the MML hierarchy using a set of seven specialized tools: functions to list accelerator systems, enumerate device families within a system, inspect available fields for a given family, retrieve channel names with optional filtering, and query accelerator physics indices.
These tools enable progressive disclosure, allowing the agent to explore systematically from broad categories (e.g., listing systems) to increasingly specific queries (e.g., retrieving filtered channel names for a particular device family and field).
Finally, in the third stage, results from parallel branches are aggregated and returned to the user as a unified response.

An important design feature is that the agent navigates exclusively through well-defined tool interfaces and cannot execute arbitrary database queries or control system commands.
This constraint ensures safe, structured exploration and prevents the agent from issuing malformed or dangerous operations.
Additionally, because all retrieved channel names must exist within the MML structure, the system naturally eliminates hallucination of non-existent process variables—a common failure mode in less constrained LLM applications.

We validated the ALS implementation on a set of 104 expert-curated queries developed in collaboration with 15 accelerator physicists.
These queries span all six accelerator systems and represent the range of channel-finding tasks encountered in routine operations.
Using Claude Haiku 4.5, the system achieves an average accuracy of 93\%.
Importantly, both cost and latency are production-viable: queries complete in seconds rather than tens of seconds, and per-query costs are a small fraction of those incurred by the earlier SQL-based approach.

\textbf{Generalization:} The ALS experience demonstrates that for legacy facilities with poor native data quality—inconsistent naming, sparse descriptions, or both—targeting an abstraction layer (such as a middle layer, SCADA interface, or other structured representation) is often the only viable path to reliable semantic channel finding.
Once such a clean abstraction exists, standard hierarchical navigation techniques can be applied effectively.
The combination of progressive tool-based exploration, dynamic example selection, and parallel query processing provides a robust framework that balances accuracy, efficiency, and safety.
This approach is readily generalizable to any facility that maintains a structured abstraction over its raw control system, regardless of the underlying naming conventions or metadata quality.
To accelerate adoption, a simplified, production-ready implementation of the middle-layer exploration pipeline is available as a plug-and-play tutorial within the Osprey framework~\cite{osprey_tutorial_channel_finder}.

\section{Paradigm 4: Ontology Approaches}\label{sec:ontology}

Direct lookup, hierarchical navigation, and interactive exploration all operate primarily over channel names and their organizational structure.
The ontology paradigm takes a complementary approach: rather than working only with how channels are named or grouped, it builds an explicit, machine-readable model of the accelerator domain itself—the devices, signals, and relationships that channels represent.
Magnets, diagnostics, power supplies, and their associated signals are represented in a structured knowledge base using semantic relationships such as ``is a kind of,'' ``part of,'' ``has setpoint,'' and ``has readback.'' Natural-language questions are interpreted in terms of these device types and relationships, then resolved via queries over the knowledge graph.
For example, ``list the setting PVs for all magnets'' is answered by following subclass relations to automatically include both quadrupoles and correctors, using explicit semantic properties rather than pattern matching over channel names.

The key advantage is portability and semantic clarity.
Because the domain model is facility-agnostic, the same query logic can be reused across laboratories: different facilities map their local naming conventions into a shared ontology, and queries written against that ontology work everywhere.
The trade-off is upfront modeling effort—constructing a domain ontology and mapping facility-specific conventions into it.
For multi-facility organizations or deployments requiring cross-site interoperability and explainability, this investment pays dividends in long-term maintainability and reusability.

\begin{figure*}
    \centering
    \includegraphics[width=0.85\linewidth]{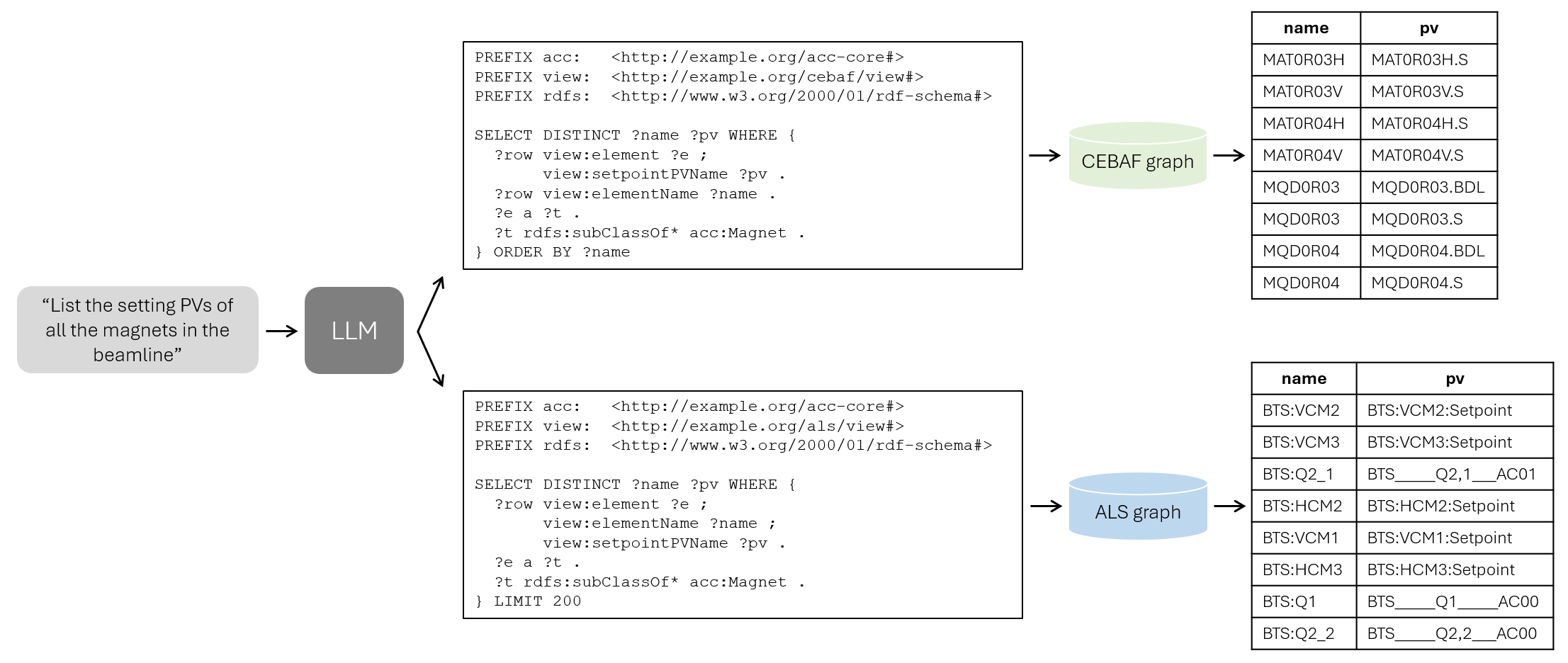}
    \caption{Example of an LLM transforming a natural language question into a formal SPARQL query. Despite the minor syntax differences, the two queries are functionally equivalent. When applied to each materialized graph, the correct PVs are returned for each facility.}
    \label{fig:ontology_schematic}
\end{figure*}

\subsection{Case Study 6: W3C Ontology Stack}\label{subsec:ontology_w3c}

A central ingredient in this case study is a domain ontology for accelerators: a small, shared vocabulary that states what kinds of things exist (magnets, diagnostics, signals) and how they relate (for example, "is subclass of," "measures," "has readback") \cite{tennant2025coreontologyparticleaccelerators}.
In many control rooms, much of this knowledge already lives implicitly in middle layers and naming conventions.
Our approach is to make a carefully chosen subset of that knowledge explicit and machine-readable, so that it can be reused across facilities and by agentic systems, not just by the local software stack.
Middle layers and property graphs remain valuable for scripting and application logic.
The ontology sits alongside them as a compact, facility-agnostic description of meaning.
In our implementation, this ontology and its instance data form a knowledge base encoded using the W3C Resource Description Framework (RDF) \cite{rdf11-concepts}, where information is stored as subject–predicate–object "triples" in a dedicated triplestore.
We then query this RDF knowledge base using the SPARQL Protocol and RDF Query Language (SPARQL) \cite{sparql11-query}, which plays a similar role for RDF that SQL plays for relational databases, but is tailored to asking questions about graph-shaped data and semantic relationships rather than tables.

To keep the design portable and to avoid inventing yet another bespoke format, the entire workflow is built on the W3C Semantic Web stack \cite{w3c}, which includes: RDF as the core data model, RDF Schema (RDFS) \cite{brickley2014rdfs} for basic typing and inheritance, the Web Ontology Language (OWL) \cite{owl2-overview} for more precise logical constraints and class relationships, and the Shapes Constraint Language (SHACL) \cite{shacl-spec} to validate that each facility's data actually conforms to the intended structure before it is exposed to tools.
Queries are written in SPARQL, which provides a standard way to retrieve "all devices of this type with these relationships" regardless of which laboratory the data came from or which triplestore is used.
By grounding the ontology, validation rules, and queries in W3C standards, we gain a technology stack that is open, well documented, and supported by multiple vendors and open-source projects.
This makes the resulting accelerator ontology more likely to remain usable and extensible over time, and easier to integrate into other data-centric and AI workflows beyond the specific case studies shown here.

Within this case study we implemented this pattern for two concrete but modest testbeds, one simple beamline segment at CEBAF and one at the ALS.
Starting from existing inventories and middle-layer exports, we first mapped each facility's local device types and PV naming conventions into a common core ontology.
For example, we assert that particular local device families are kinds of Quadrupole or Beam Position Monitor, and that certain PVs play the role of setpoints or readbacks.
These mappings are used to generate materialized RDF graphs that instantiate the core concepts with actual devices and channels from each site.
The resulting CEBAF and ALS graphs are loaded into standards-compliant triplestores and exposed as SPARQL endpoints.
In effect, we create two different "views" of two different machines that nonetheless speak the same semantic language, so that a single query template over the ontology can be reused at both facilities.

On top of these graphs, we add a thin natural-language interface using a local language model.
The model is given a compact description of the ontology and a handful of examples showing how typical operator questions correspond to SPARQL patterns.
Its role in the overall agentic system is deliberately narrow: it translates operator intent into structured SPARQL queries against the RDF knowledge base (see Figure~\ref{fig:ontology_schematic}).
The query results are channel names and associated metadata, which can then be handed off to other tools in the agentic stack (archive readers, plotting tools, analysis routines).
This separation of concerns is key.
The ontology and RDF graph provide a stable, auditable substrate for meaning, while the language model and agents focus on interpretation and orchestration.
Together, they turn the challenge of finding relevant channels from a site-specific craft into a reusable capability that can travel with the agentic framework from one facility to another.

In this work, the ontology-based implementations at CEBAF and ALS are used primarily to demonstrate feasibility and design patterns rather than to present a large-scale quantitative benchmark.
The focus is on showing that a compact, facility-agnostic accelerator ontology, combined with materialized RDF graphs, enables a single SPARQL query template to travel across facilities and be driven safely by a narrow LLM-to-SPARQL translation layer.
Extending this paradigm to larger, shared validation query sets and more detailed performance studies is a natural next step.

\textbf{Generalization:} The ontology-based pattern developed in this case study is applicable to any facility that can (i) define a compact domain ontology for its key accelerator components and signals, (ii) map local device catalogs and PV naming conventions into that shared vocabulary, and (iii) expose the resulting knowledge base via a standards-compliant SPARQL endpoint.
Once these prerequisites are met, a narrow natural-language-to-SPARQL translation layer, together with a small library of query templates, can be reused across machines and even laboratories, allowing semantic channel-finding capabilities to travel with the agentic framework rather than being re-engineered for each site.
Although our CEBAF and ALS implementations currently target modest beamline segments, the underlying W3C stack and modeling approach are designed to scale to larger machines and additional subsystems as ontological coverage grows.
More broadly, the same architecture can support semantic channel finding in other complex experimental infrastructures—such as synchrotron light sources, FELs, and fusion devices—whenever they adopt a similar ontology-backed knowledge base over their control system.

\section{Future Directions}\label{sec:future}

The four paradigms presented in this work—direct lookup, hierarchical navigation, interactive exploration, and ontology-based approaches—provide a practical foundation for semantic channel finding in current control systems.
At the same time, several extensions and enhancements emerge naturally from our implementations.

A first direction is to explore hybrid approaches that combine multiple paradigms adaptively based on query characteristics and available infrastructure.
For example, ontology reasoning could inform hierarchical navigation strategies, while semantic embeddings could refine direct lookup in ambiguous cases.
Such hybrids would allow facilities to exploit existing structure where it is available, while falling back to lightweight in-context methods for small, well-described channel subsets.

A second direction concerns dynamic long-term memory for ReAct-style agents.
Our current implementations rely on static few-shot examples selected at query time, as in the ALS PV finder.
Future systems could instead maintain a persistent memory of successful query patterns, operator corrections, and facility-specific navigation strategies.
This would allow agents to learn from operational deployment, capturing frequently asked queries, common error modes, and emerging domain-specific terminology.
Shared and continuously updated memory across agent instances could improve performance over time without ongoing manual curation.
The key technical challenge is obtaining reliable feedback signals when channel finding is embedded in larger workflows, and attributing task success or failure to specific navigation paths when the ultimate outcome may occur several steps downstream of the initial lookup.

A third natural extension is to turn semantic channel finding into a driver for systematic data-quality improvement.
Across all four paradigms, failures and operator corrections provide rich signals about ambiguous names, missing descriptions, and underspecified metadata.
Future systems could track these patterns over time, propose edits to channel descriptions or middle-layer abstractions, and surface conflicts to maintainers for review.
This would create a feedback loop in which semantic channel-finding infrastructure not only consumes existing naming conventions and metadata, but also helps to gradually repair and standardize them.

Reinforcement learning for tool-using agents offers another promising direction.
At present, ReAct-style agents use large models in few-shot mode; RL provides a path to smaller, faster, facility-specific models.
The validation query sets developed in this work provide natural reward signals for training such agents.
Modern RL techniques require relatively few examples to learn effective tool-selection strategies, making this approach viable even for facilities with modest curated query sets (on the order of 100 queries).
A reusable training pipeline in which facilities fine-tune open-source language models to their local database schema and query patterns could yield higher reliability, lower latency, and reduced dependence on cloud APIs.
The main technical difficulty lies in combining in-GPU LLM training with external tool-calling environments, an area where we expect rapid progress as agentic workloads become more widespread.

Finally, ontology-based methods open the door to cross-facility knowledge transfer.
Because ontologies provide a facility-agnostic representation of devices, signals, and relationships, navigation strategies learned at one machine can in principle bootstrap channel finding at similar facilities.
Extending the accelerator ontology to cover a broader range of facility types and subsystems, and developing transfer-learning strategies that reuse query templates and reasoning patterns across sites, are natural next steps.

Taken together, these directions outline a path from the prototype systems described here to mature, facility-integrated assistants that combine language models, structured control-system abstractions, and cross-facility semantics.
Although our case studies focus on accelerator control rooms, the same patterns apply whenever complex experimental infrastructures must be exposed to human operators and AI agents through robust, auditable language interfaces.

\section*{Acknowledgments}

We acknowledge support from DESY (Hamburg, Germany), a member of the Helmholtz Association HGF and European XFEL GmbH (Schenefeld, Germany). Part of this research was enabled by the Maxwell computational resources operated at Deutsches Elektronen-Synchrotron DESY, Hamburg, Germany.

This work was supported by the Director of the Office of Science of the U.S.~Department of Energy under Contract No. DE-AC02-05CH11231 and U.S. Department of Energy, Office of Science, Office of Nuclear Physics under Contract No. DE-AC05-06OR23177.


\bibliographystyle{unsrt}
\bibliography{references}

\end{document}